\documentclass{article}
\usepackage{spconf,amsmath,graphicx}
\usepackage[hyphens]{url}
\usepackage{color}
\usepackage{comment}
\usepackage{floatrow}

\def\ie{{\em i.e.}}
\def\eg{{\em e.g.}}
\def\etal{{\em et al. }}
\title{De-identification without losing faces}
\name{Yuezun Li ~~~  ~~~ Siwei Lyu}
\address{Computer Science Department \\ University at Albany, State University of New York, USA}
\begin{document}
%
\maketitle
\begin{abstract}
Training of deep learning models for computer vision requires large image or video datasets from real world. Often, in collecting such datasets, we need to protect the privacy of the people captured in the images or videos, while still preserve the useful attributes such as facial expressions. In this work, we describe a new face de-identification method that can preserve essential facial attributes in the faces while concealing the identities. Our method takes advantage of the recent advances in face attribute transfer models, while maintaining a high visual quality. Instead of changing factors of the original faces or synthesizing faces completely, our method use a trained facial attribute transfer model to map non-identity related facial attributes to the face of donors, who are a small number (usually 2 to 3) of consented subjects. Using the donors' faces ensures that the natural appearance of the synthesized faces, while ensuring the identity of the synthesized faces are changed. On the other hand, the FATM blends the donors' facial attributes to those of the original faces to diversify the appearance of the synthesized faces. Experimental results on several sets of images and videos demonstrate the effectiveness of our face de-ID algorithm. 
\end{abstract}
%

\section{Introduction}
\label{sec:intro}

\begin{figure*}[t]
\centering
\includegraphics[width=1\linewidth]{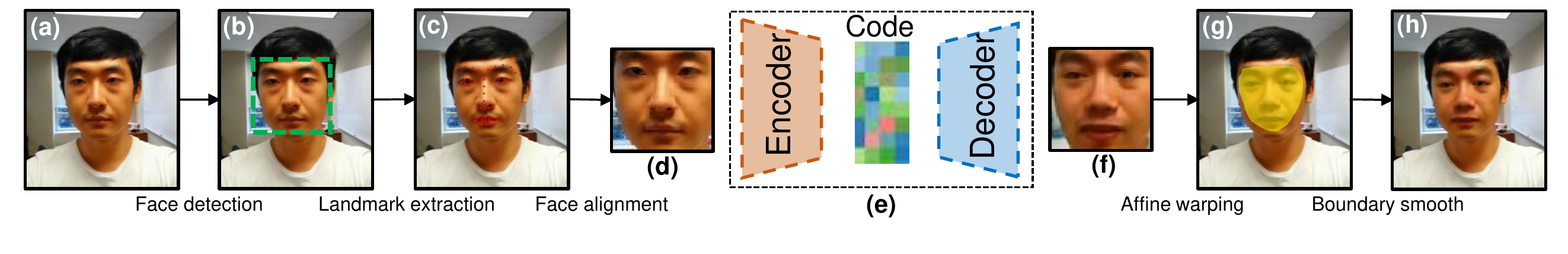}  
\vspace{-1cm}
\caption{\small \em Overview of face de-identification. (a) is the original input image. The green dash box in (b) is the detected face area. Then facial landmarks are extracted in (c), marked by red points. (d) is the aligned face area, which is the input of face synthesizer (e). The architecture details of face synthesizer is illustrated in (j), where the blue masked area is the test mode used in (e). (f) is the synthesized face area, which is then affine warped back to original image as (g). To remove the artifacts, we only retain face area inside the mask, which is made based on facial landmarks. Finally, we smooth the boundary to further reduce the artifacts.}
\label{fig:overview}
\end{figure*}

Recent years have seen great successes of deep neural networks in solving various computer vision problems including face detection, face recognition and emotion classification. The training of deep neural networks predicate on large-scale and carefully annotated image/video datasets. However, unlike images and videos enacted by consented subjects, for those collected from real world, the law requires that the privacy of the people inadvertently captured by camera need to be protected before such data can be used. As face is the most identifiable part of a human, visual anonymity can be achieved by changing the faces, a problem commonly known as {\em  face de-identification} (face de-ID). 

The simplest face de-ID method is to obfuscate faces in images by blurring or pixelation (\eg, in Google Map Street View). However, it is not as effective as one may think, because it is possible to identify a particular subject by comparing faces after the obfuscation operations, known as a {\em parrot attack} \cite{newton2005preserving}. Moreover, the complete removal of faces from images and videos makes them useless for training deep neural networks that analyze facial expressions or other non-identity related attributes. Moreover, images and videos with faces obfuscated do not look ``natural''.  

More sophisticated face de-ID methods focus on changing faces rather than removing them. Early works (\eg, \cite{newton2005preserving,gross2005integrating,gross2006model,gross2008semi}) generate de-IDed faces by removing high frequency details, but they usually lead to faces with blurred appearances. The developments of image synthesis methods based on deep neural networks, in particular, generative adversary networks (GANs) \cite{goodfellow2014generative}, inspire a new vein of face de-ID methods \cite{brkic2017know,sun2018natural}, which uses synthesized faces to replace the originals. However, these methods typically requires a large number of face images in training. Furthermore, they cannot be extended to face de-ID tasks for videos, as they can only generate individual images and cannot maintain temporal consistency between video frames. 

In this work, we describe a new face de-ID method based on a deep neural network based image style transfer model \cite{NIPS2017_6672}. Our method treats the non-identity related facial attributes as the style of the original face, and use a trained facial attribute transfer model (FATM) to map them to the face of donors, who are a small number (usually 2 to 3) of consented subjects. The FATM  is composed of an encoder and a decoder. The encoder maps the input original face to an identity-neural representation (the code), and the decoder combines the code with the donors' identity to create new faces. Using the donors' faces ensures the natural appearance of the synthesized faces. On the other hand, FATM blends the donors' facial attributes to those of the targets' to generate synthesized faces of different identities. The training of FATM can be achieved with much smaller set of images -- typically $\sim 500$ images is enough in comparison to tens of thousands required to train a full blown GAN model. This means efficient training and run-time efficiency. Experimental results on several sets of images and videos demonstrate the effectiveness of our method.   

\section{Related works}
\label{sec:format}


Early methods,\eg, \cite{newton2005preserving,gross2005integrating,gross2006model,gross2008semi}, substitute original faces with the average of face images of the $K$-closest identities to the subject from a closed set of facial images. Subsequently, variations in face poses are considered to improve the robustness of face de-ID methods in  \cite{samarzija2014approach}. {The work \cite{bhattarai2014puzzling} de-identified the face images by adding designed noise patterns.} In \cite{jourabloo2015attribute}, a new objective function combining face de-ID and face verification is introduced to ensure the original and de-IDed face to have common facial attributes but different identities. {The work of \cite{letournel2015face} proposed an adaptive filtering method for face de-identification with expressions preserved in images.} The diversity of the de-IDed faces is considered in \cite{sun2015distinguishable} to avoid generating faces that all look alike.

More recently, deep neural networks have been used for face de-ID. The work of \cite{meden2017face} uses GANs  to generate de-ID faces, which is extended by Karla \etal in \cite{brkic2017know} for full body synthesis. However, the GAN synthesized de-IDed faces suffer from artifacts such as the skin color disparity between the de-IDed face and the surrounding area. Original faces are partially replaced in \cite{sun2018natural} using GAN based in-painting method, which uses facial landmarks as an input to the GAN model for consistent head poses with the original faces. However, temporal consistencies of faces across different video frames and subtle face attributes are not well preserved in this method.

\section{Methods}

In this work, we describe a different approach to face de-ID based on the neural network based image style transfer model of \cite{NIPS2017_6672}. We use synthesized faces created by transferring facial expressions of the original subject (the 'target') to the faces of another subject (the 'donor'), a consented subject who grants rights to use his/her face images. The replacement of the target's facial attributes with the donor's conceal the target's identity, while the preserved facial expressions keep the utility of the resulting image as training data. 

\begin{figure}[t]
\centering
\includegraphics[width=1\linewidth]{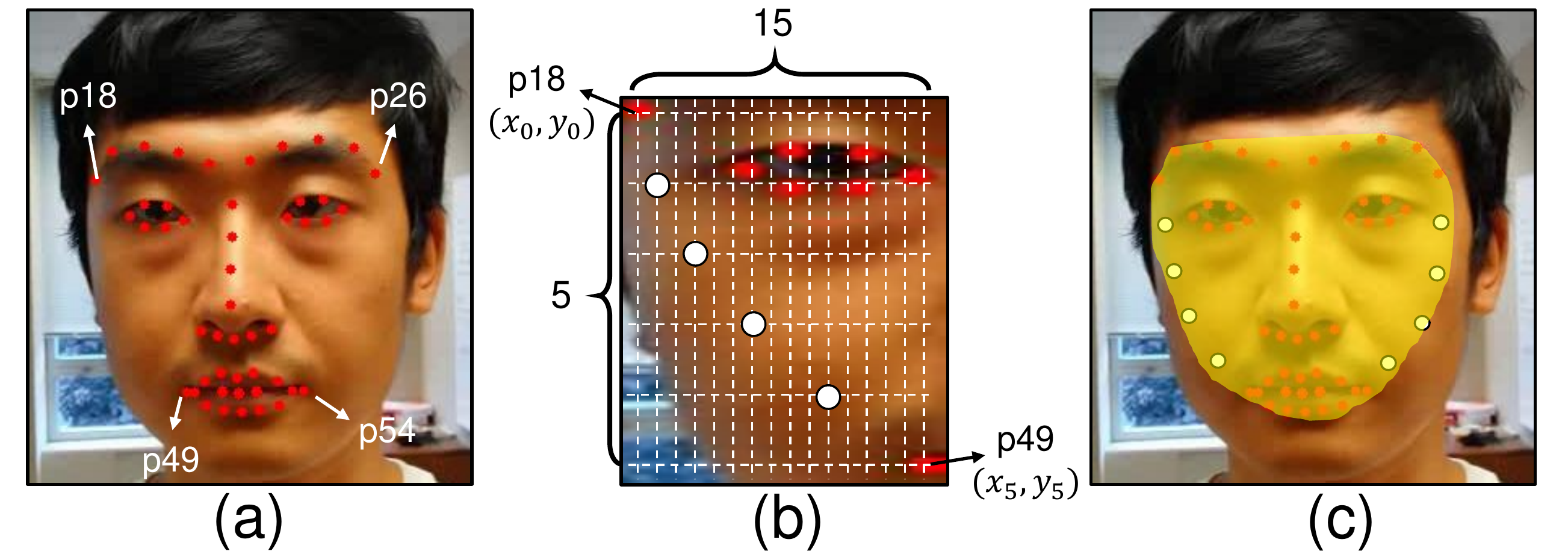}
~\vspace{-2em}
\caption{\small \em Creating face mask. (a): original image with facial landmarks, (b): interpolated points on the boundary of the mask (for left face), (c): final face mask (in yellow) using landmark points and their interpolations.}
\label{fig:mask}
~\vspace{-2em}
\end{figure}

The overall pipeline of our face de-ID method is shown in Figure \ref{fig:overview}.  The input is a RGB image or video frame containing the face of the target.  We first run a face detector and crop each detected face using the bounding boxes. Then, a facial landmark extraction algorithm is applied to the extracted face to locate landmark points corresponding to distinct facial structures such as the tips of eyes, eyebrows, nose, mouth and contour. These landmark points are then matched to the landmark points of a ``standard'' face, which has a fixed size with a frontal orientation, with an affine transform. The affine transform is obtained by minimizing the distortion between the two sets of landmark points. Using this affine transform, we then warp every pixel of the extracted face to the pose of the standard face, and resize it to have dimension of $64 \times 64$ pixels.  The rectified face is fed to the facial attributes transfer model (FATM), which will be described in detail subsequently. FATM synthesizes a face based on the donor's identity and the facial expression, head orientation, lighting condition, skin color and other facial characteristics of the target's face. The synthesized face image is resized to the original face, and warped back to the original configuration using the inverse of the same affine transform previously estimated. After that, the synthesized face is trimmed with a face mask obtained from the landmark points to blend into the surrounding context. The face mask is created from the convex hull of landmarks of the eye browns and the bottom outline of mouth, and $8$ interpolated points on both side of the faces to maximally cover the facial area. For instance, from the left side of the face, we choose two extreme landmark points corresponding to the leftmost tip of the eyebrow and leftmost tip of the mouth, the coordinates of which are denoted as $(x_0, y_0)$ and $(x_5, y_5)$, respectively. Then we use an interpolation scheme to generate four more points in between these two landmark points $x_i = x_{i-1} + \frac{i}{15}(x_5 - x_0),y_i = y_{0} + \frac{i}{5}(y_5 - y_0)$ for $i=1,2,3,4$, Figure \ref{fig:mask}. A similar procedure is repeated for the right side of the face. As the last step, we apply adaptive Gaussian smoothing of the boundary before finally splicing it into the original image to conceal the boundary of splicing. The whole process is automated and runs with minimum manual intervention.

\subsection{Facial Attribute Transfer Model (FATM)} 

\begin{figure}[t]
\centering
\includegraphics[width=0.95\linewidth]{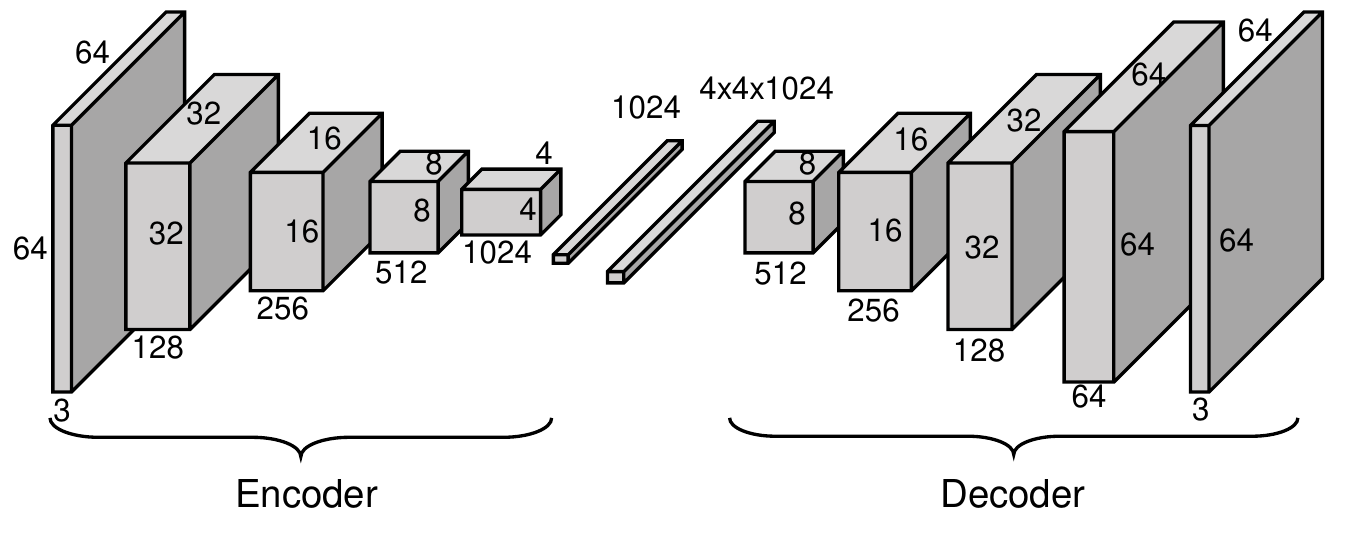}  
\vspace{-0em}
\caption{\small \em The neural network architecture of the facial attribute transfer model (FATM).}
\label{fig:network}
\vspace{-2em}
\end{figure}

The facial attribute transfer model (FATM) is the core component of our face de-ID method. Inspired by the deep image style transfer framework \cite{NIPS2017_6672}, FATM is composed by a pair of deep neural networks: the encoder and the decoder. The encoder converts the input face to a representative feature (the 'code'), and the decoder reverses the process to synthesize a face from the code. Specifically, we refer to face images of the same subject as a {\em face set}. Different face sets share the same encoder $E$, but each have a dedicated decoder. This specific structure is to ensure the encoder to capture the identity-independent attributes common to all face sets, while the individual decoders can preserve identity-dependent attributes of each subject and map such attributes onto the synthesized faces. 

The specific neural network architecture of the encoder and the decoder is shown in Figure \ref{fig:network}. The encoder has four convolution (Conv) layers and two fully connected (FC) layers. The four convolution layer has $128,256,512$, and $1024$ convolution kernels, respectively. The convolution kernels all have size $5 \times 5$ pixels with stride of $2 \times 2$ pixels. The leaky RELU function, defined as $f(x)=\max(0.1x, x)$, where $x$ is the input, is adopted as the nonlinear activation function of each convolution layer. The two fully connected layers have dimensions $1,024$ and $16,384$, respectively. The code is the output of the last fully connected layer in the encoder, which is a $16,384$-dimensional vector. Similarly, the decoder has four de-convolution (Upscale) layers, with $512, 256, 128$, and $64$ convolution kernels of size $3 \times 3$ and strides $1 \times 1$ pixels, respectively. The nonlinear activation function for these convolution layers is the same leaky RELU function as in the encoder. The final output from the decoder is reshuffled to 2D images of $64 \times 64$ pixels, and the final synthesized face of RGB color is produced using $3$ convolution kernels of size $5 \times 5$ with stride $1$ on last layer. 

\begin{figure}[t]
\centering
\includegraphics[width=0.95\linewidth]{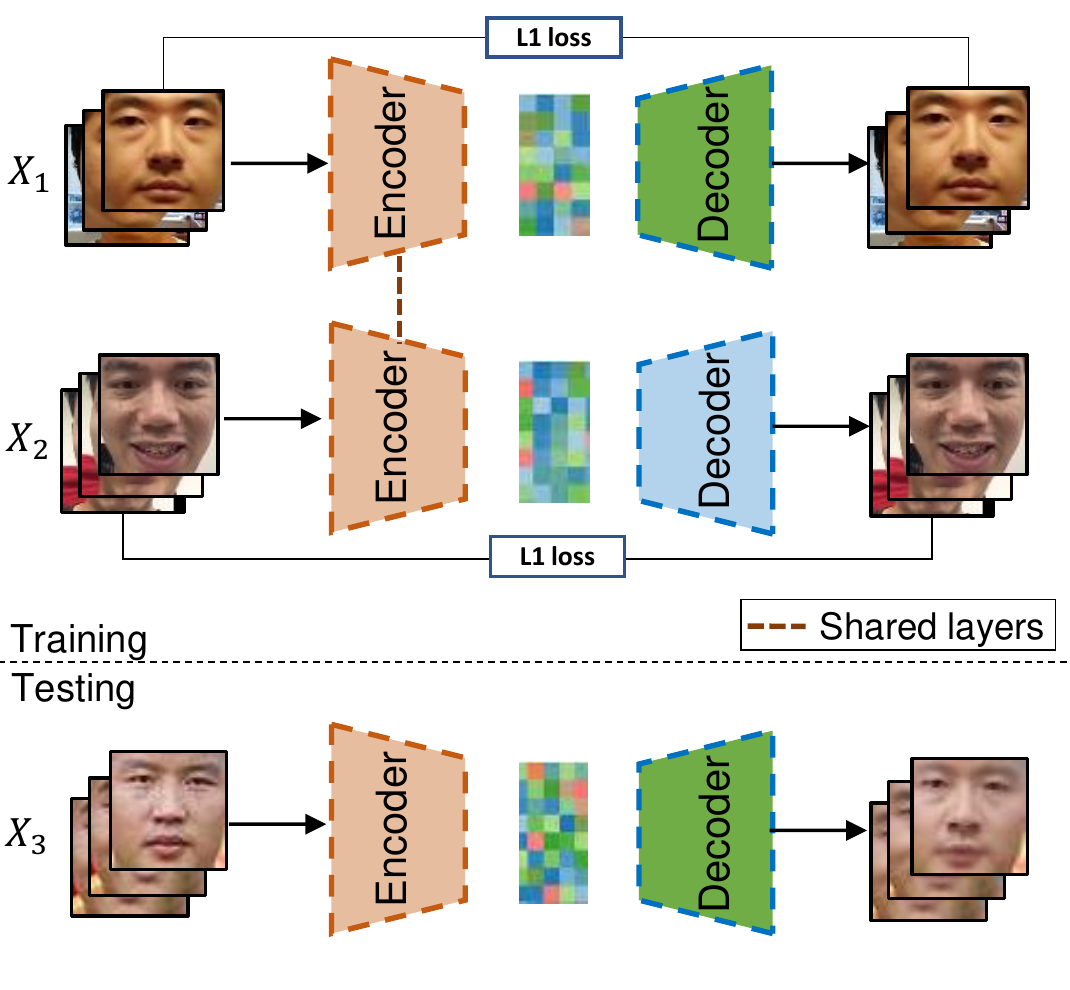}  
\vspace{-1em}
\caption{\small \em Training and deployment of FATM.}
\label{fig:train}
~\vspace{-2em}
\end{figure}

\subsection{Training FATM}

The encoder and decoder networks are trained in tandem in an unsupervised manner, using face sets from multiple subjects but do not need to have any correspondence in facial attributes such as expressions, head orientations, lighting, etc, so relatively little labeling effort is required. The face sets are first processed with face detection, landmark extraction and rectification to be the training data for the two networks. 

Learning FATM is equivalent to find optimal parameters for the common encoder $E$, and individual decoder $D_i$. Figure \ref{fig:train} illustrate the training of the FATM with two face sets $X_1$ and $X_2$. Specifically, we first use $E$ and $D_1$ to form an encoder-decoder pair, and optimize their parameters to minimize the reconstruction errors for faces in $X_1$. The reconstruction error for one face $\vec{x} \in X_1$ is given by $\|\vec{x}-D_1(E(\vec{x}))\|_1$. The parameter update is performed with the back-propagation algorithm implemented with stochastic gradient update with an ADAM optimizer. We set the batch size to $64$, and the initial learning rate to be $5 \times 10^{-5}$. Then, a similar procedure is performed for $X_2$, this time with the encoder-decoder pair $E-D_2$. When updating on $X_2$ is complete, we go back updating the parameters of encoder-decoder pair $E-D_1$ and the iteration goes on for $10^6$ times. 

To improve the visual quality of the synthesized faces, we also take several measures to increase the diversity of the training data.  In each training round, we use input face regions that are slightly larger than $64 \times 64$, and then select randomly cropped $64 \times 64$ face regions iteration to simulate the variations of locations of faces; we also apply random rotation, horizontal mirroring, and scaling to the faces to simulate different viewing angle and distance of the faces. Variations in skin color affect the visual quality of generated faces and the major cause of conspicuous artifact in the synthesized faces. Hence, we further randomize the color of the training faces in the brightness, contrast, distortion and sharpness in each iteration to simulate the variations in skin color. 

\section{Evaluations}
\label{sec:exp}

We perform several sets of experiments to evaluate the performance of our face de-ID algorithm and compare with state-of-the-art methods. 

\smallskip
\noindent{\bf Datasets:} We use donor faces from six individuals who have signed consensus forms for the use of their face images. The donor face set is obtained from 60 video clips (10 from each of the six subjects) of approximate 30 seconds in length (30 frame-per-second) and $1920 \times 1080$ or $1280 \times 720$ pixels in resolution. As a result, we have in total $540,000$ high resolution face images to train the FATM model. 

We evaluate our method using two popular face image datasets, namely the LFW dataset \cite{learned2016labeled} and the PIPA dataset \cite{zhang2015beyond}. The LFW dataset is designed for testing face verification performance. As such it contains around $13,000$ images of faces collected from the Internet. The size of image in LFW is fixed to $250 \times 250$ pixels. PIPA dataset \cite{zhang2015beyond} is a more challenging dataset, which contains $37,107$ images collected from public Flickr photo albums in an unconstrained setting. This dataset has about $2,000$ individuals with diverse poses, clothing, camera viewpoints, lighting conditions and image resolutions. 

\smallskip 
\noindent{\bf Runtime details}. We use the face detection and landmark location functionalities from package {\tt DLib} \cite{Dlib09}. The training and evaluation of our algorithm is performed on a computer with Intel Xeon(R) CPU X5570 \@ 2.93GHz and NVIDIA GTX GPU. The code implementing the training and evaluation uses Google Tensorflow 1.3.0 with CUDA 8.0 on Ubuntu 16.04.  The training time for FATM is around {72 hours} on our current training dataset. Generating a synthetic de-IDed face including post-processing takes about $0.24$ seconds on average. 

\begin{figure*}[t]
\centering
\includegraphics[width=0.9\linewidth]{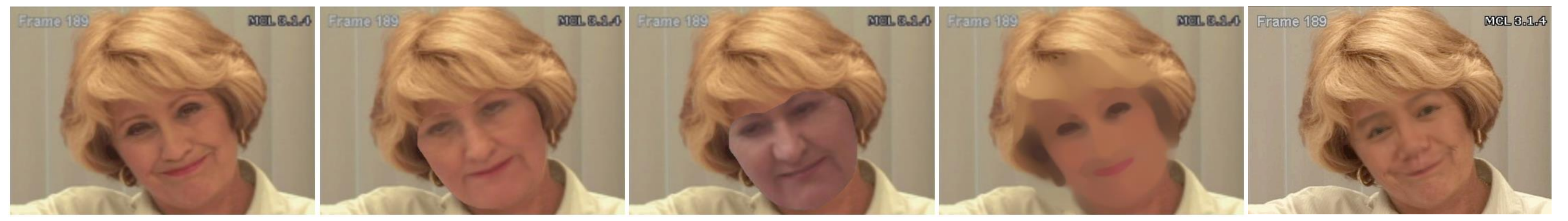}
\par \hspace{1.5em} Original \hspace{4em} $k$-Same \cite{newton2005preserving} \hspace{3.5em} MF($\epsilon, k$) \cite{gross2009face} \hspace{1.5em} Adaptive filtering  \cite{letournel2015face} \hspace{1.9em} Our method
\caption{\small \em The comparison of visual quality of different face de-ID methods.}
\vspace{-0.5cm}
\label{fig:compare}
\end{figure*}

\begin{figure}[t]
\centering
\includegraphics[width=\linewidth]{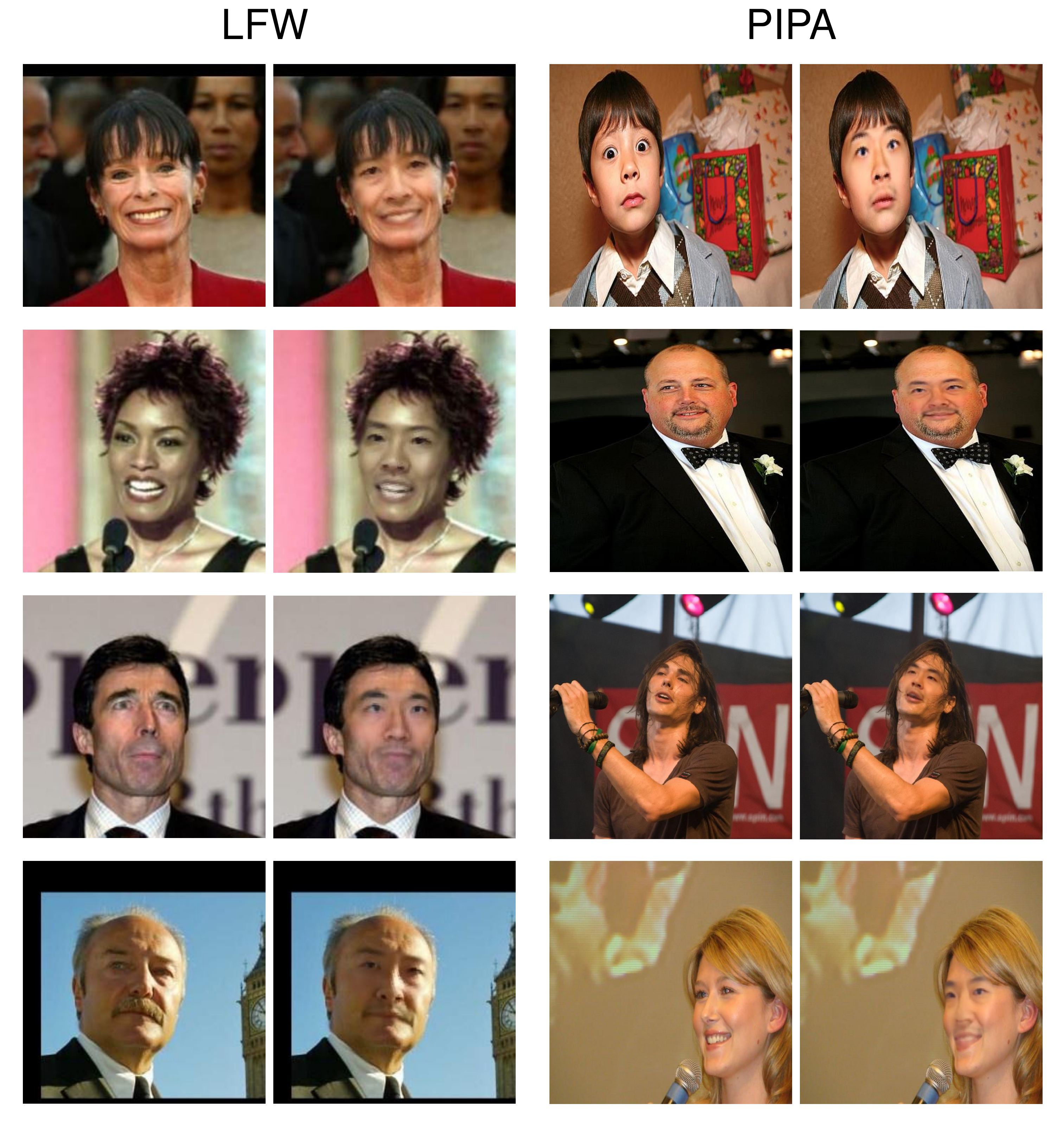} 
\vspace{-2em}
\caption{\small \em Examples of face de-ID for images from LFW and PIPA. The left image is the original while the right is the de-IDed face. }
\label{fig:exps}
\vspace{-2em}
\end{figure}

\begin{figure}[t]
\centering
\includegraphics[width=\linewidth]{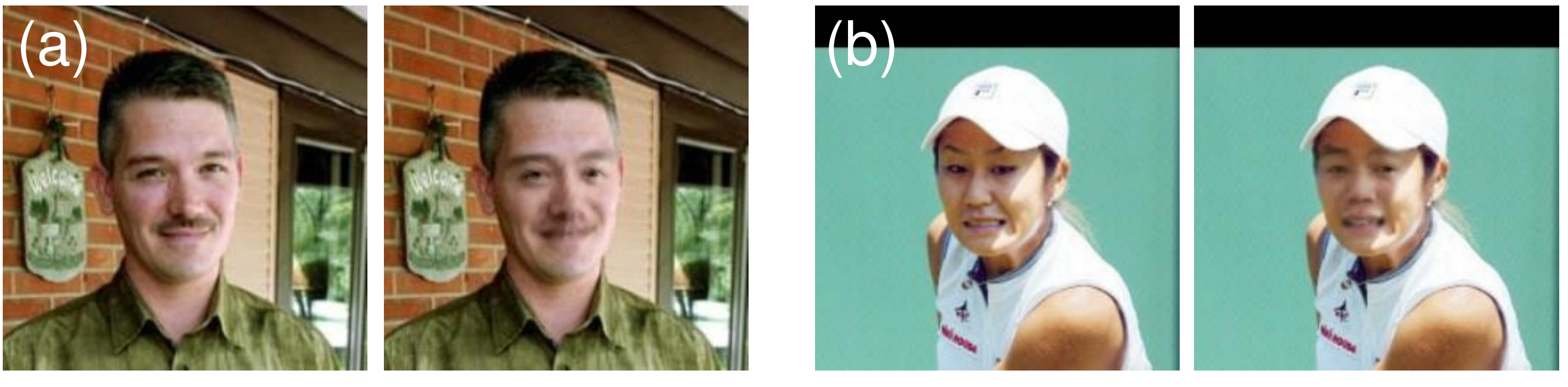} 
\vspace{-0.8cm}
\caption{\small \em {Examples that are visually different, but are determined as from a same subject by face verification algorithm.}}
\label{fig:fail-case}
~\vspace{-2em}
\end{figure}

\begin{figure}[t]
\centering
\includegraphics[width=\linewidth]{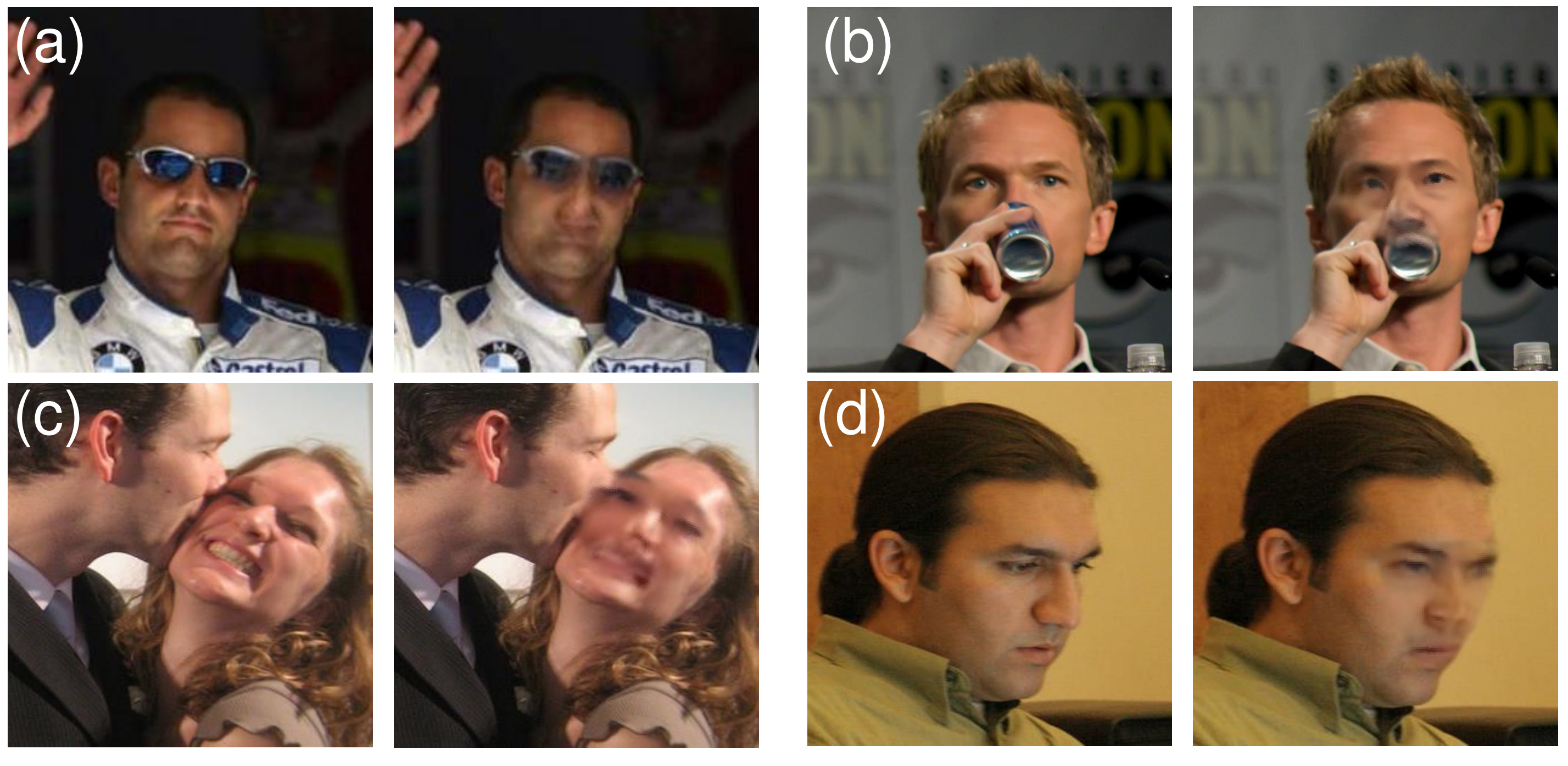} 
\vspace{-0.5cm}
\caption{\small \em Examples of failure cases. (a,b) are cases that face is occluded by other objects. (c) is the uncommon facial expression. (d) is the strongly non-frontal head orientation.}
\label{fig:fail-case-2}
~\vspace{-3em}
\end{figure}

\begin{figure*}[t]
\centering
\includegraphics[width=0.95\linewidth]{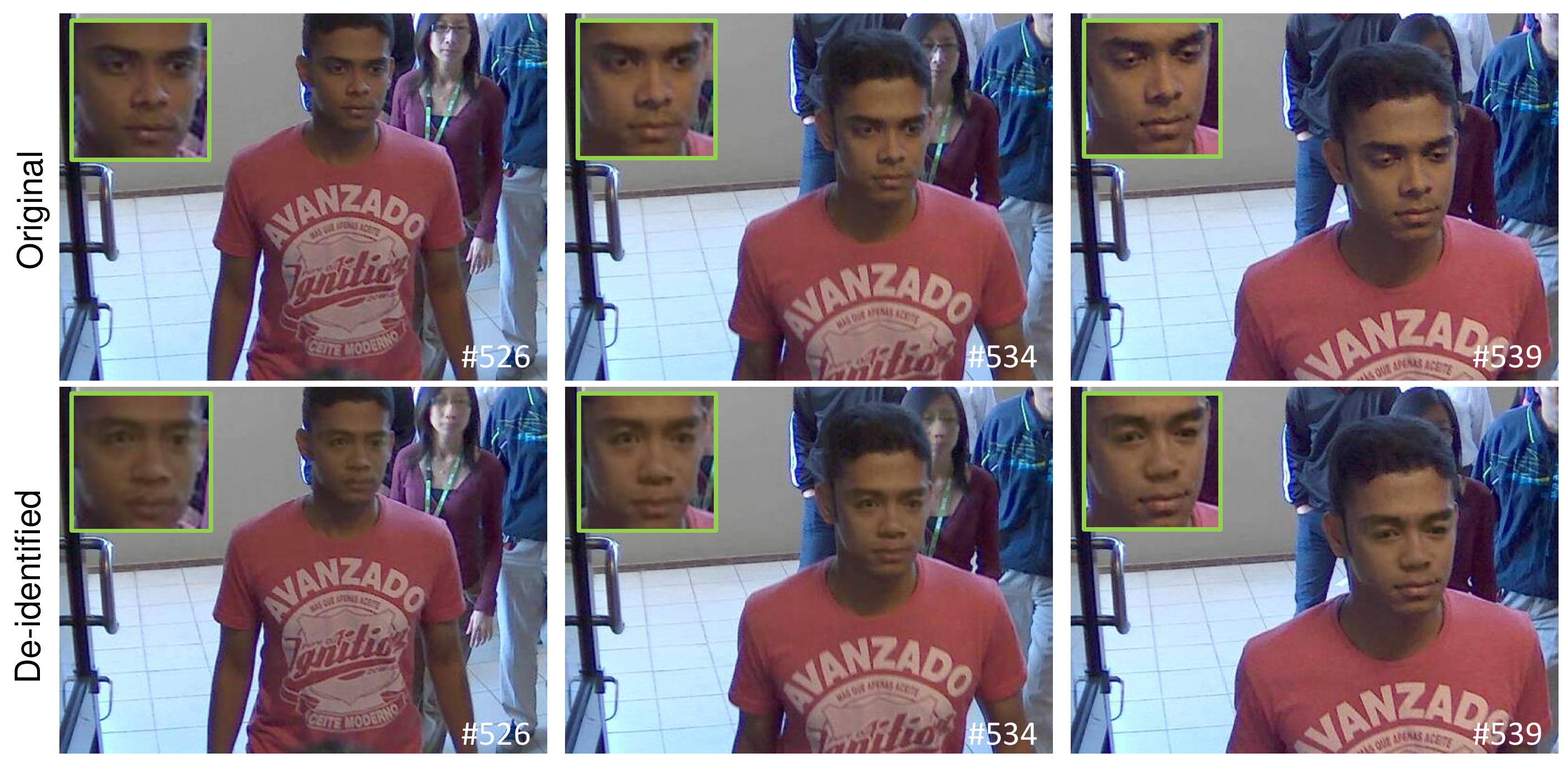}
~\vspace{-1em}
\caption{\small \em Illustration of de-identification on video sequence. The first row is original video and the second row is de-identified video. The green box is zoomed in face area for better visualization.}
\label{fig:video-demo}
~\vspace{-2em}
\end{figure*}

\smallskip
\noindent{\bf Evaluating face de-identification}. {To provide a quantitative performance evaluation of the face de-identification, we follow the work of \cite{letournel2015face} that uses face verification evaluation on the LFW dataset for this purpose. Specifically, we randomly select $1,000$ image pairs from the LFW dataset, each corresponding to two images of the same subject differing in background, head pose, apparels and/or facial expressions. We apply our face de-ID method on one image in each pair and then feed both images to a state-of-the-art face verification algorithm provided by {\tt Dlib}\footnote{The {\tt Dlib} face verification algorithm is based on the ResNet-34 network \cite{he2016deep} and can achieve $99.38\%$ accuracy on LFW dataset.} to determine if they are from the same subject. If the de-identification is effective, the two images should be classified as from different identities. 
On the $1,000$ pairs, the face verification accuracy is $97.6\%$ and $16.5\%$ before and after de-identification respectively, \ie, $83.1\%$ are determined to be from different subjects. In comparison, the method of \cite{letournel2015face} is only $34.0\%$ effective in de-identifying the subjects. } 

We also conduct a {\em self de-identification} experiment \cite{bhattarai2014puzzling}, where we compare the de-IDed image with its corresponding original image. In this case, all other factors stay the same and the only change to each image occur at the face region. However, in this case, the effective rate of de-identification drops to $67.2\%$. In particular, as shown in Figure \ref{fig:fail-case}, even though many de-IDed images visually appear to be from different subject, the face verification algorithm determines they are from the same subject nevertheless.  This is a puzzling result, but we speculate that it is due to the specific design of face verification algorithm. Specifically, our method only replaces the center area of the face, and leaves the target's hair and face shape unchanged. However, hair and face shape are two cues for the {\tt Dlib} face verification algorithm, so some of such faces are still being classified as from the same subject, even though the locations of facial parts are different.

\smallskip
\noindent{\bf Comparing visual qualities.} We show several examples of the de-IDed images in Figure \ref{fig:exps} using images from the LFW dataset and the PIPA dataset, respectively. One potential limitation of our method is that we only use limited number of donors, which may reduce the diversity of the synthesized de-IDed faces. However, visual examples of de-IDed faces as shown in Figure \ref{fig:exps} suggest that this is not the case. We think the reason is that the learned decoder in the FATM model is capable of mixing facial attributes of the target with those of the donor, and in doing so creates new face images with variations in skin color, facial characteristics and expressions that are different from the original donors. This further improves the naturalness of the de-IDed faces. Figure \ref{fig:video-demo} shows an example of our method on a surveillance video from the ChokePoint  dataset \cite{wong2011patch}. Note that the replacement of central face area in our method results in better temporal consistencies.

Figure \ref{fig:compare} shows a comparison of the visual quality of our method with that of several previous face de-ID methods including the $k$-Same method \cite{newton2005preserving}, MF($\epsilon, k$) \cite{gross2009face}, and adaptive filtering \cite{letournel2015face}. As we see from the results, other face de-ID methods introduce various artifacts, such as blurring and change of facial expressions. In comparison, our method exhibits better visual quality and the original facial expression.    

{To quantitatively analyze the visual quality, we randomly select $1,000$ images from LFW and PIPA dataset respectively and run our algorithm over them. We evaluate the visual quality of de-IDed images using SSIM \cite{wang2004image}. The higher SSIM score denotes the better visual quality. The average SSIM scores for our method are $0.97$ on LFW and $0.96$ on PIPA. In comparison, the most recent work \cite{sun2018natural} has an average $0.90$ SSIM score on PIPA.}

\smallskip
\noindent{\bf Failure Cases:} However, there are also cases when the neural network based FATM fails to generate a good face image, as shown in a few examples in Figure \ref{fig:fail-case-2}.  The failures can be attributed to occlusions of the target face by other objects (\eg, eye glasses), unusual facial expressions, and strongly non-frontal head orientations. 

\section{Conclusion}
In this work, we describe a new face de-identification method that can preserve essential facial attributes in the faces while concealing the identities. Our method takes advantage of the recent advances in face attribute transfer models, while maintaining a high visual quality. Instead of changing factors of the original faces or synthesizing faces completely, our method use a trained facial attribute transfer model to map non-identity related facial attributes to the face of donors, who are a small number (usually 2 to 3) of consented subjects. Using the donors' faces ensures that the natural appearance of the synthesized faces, while ensuring the identity of the synthesized faces are changed. On the other hand, the FATM blends the donors' facial attributes to those of the original faces to diversify the appearance of the synthesized faces. Experimental results on several sets of images and videos demonstrate the effectiveness of our face de-ID algorithm.

For future works, we would like to improve the neural network based FATM to handle more variations in head poses, lighting and facial occlusions. Furthermore, randomness can be introduced to the synthesize process to improve the diversity of the faces and remove the original target's identity more effectively. 

\smallskip
\noindent{\bf Acknowledgement}. This material is based upon work supported by the United States Air Force Research Laboratory (AFRL) and the Defense Advanced Research Projects Agency (DARPA) under Contract No. FA8750-16-C-0166. The views, opinions and/or findings expressed are those of the author and should not be interpreted as representing the official views or policies of the Department of Defense or the U.S. Government.

{\small
\bibliographystyle{IEEEbib}
\bibliography{refs}
}
\end{document}